\documentclass[sigconf]{acmart}

\AtBeginDocument{%
  \providecommand\BibTeX{{%
    \normalfont B\kern-0.5em{\scshape i\kern-0.25em b}\kern-0.8em\TeX}}}

\copyrightyear{2022}
\acmYear{2022}
\setcopyright{acmcopyright}
\acmConference[KDD '22]{Proceedings of the 28th ACM SIGKDD Conference on Knowledge Discovery and Data Mining}{August 14--18, 2022}{Washington, DC, USA}
\acmBooktitle{Proceedings of the 28th ACM SIGKDD Conference on Knowledge Discovery and Data Mining (KDD '22), August 14--18, 2022, Washington, DC, USA}
\acmPrice{15.00}
\acmDOI{10.1145/3534678.3539280}
\acmISBN{978-1-4503-9385-0/22/08}

\usepackage{amsthm}

\usepackage{amsfonts,amssymb}
\usepackage{amsmath}
\usepackage{bm}
\usepackage{multirow}
\usepackage{graphicx}
\usepackage{enumitem}
\newtheorem{myDef}{\textbf{Definition}}
\newtheorem{myProb}{\textbf{Problem}}
\usepackage[table,xcdraw]{xcolor}
\usepackage{booktabs}

\begin{document}
\renewcommand{\shortauthors}{Bin Lu et al.}

\begin{sloppypar}
\title{\textsc{Geometer}: Graph Few-Shot Class-Incremental Learning via Prototype Representation}


\author{Bin Lu, Xiaoying Gan{$^{*}$}, Lina Yang, Weinan Zhang, Luoyi Fu, Xinbing Wang}
\affiliation{
  \institution{Shanghai Jiao Tong University}
  \city{Shanghai}
  \country{China}
}
\email{{robinlu1209, ganxiaoying, alina_yln, wnzhang, yiluofu, xwang8}@sjtu.edu.cn}








\renewcommand{\shortauthors}{Bin Lu et al.}
\thanks{Xiaoying Gan is the corresponding author.}
\begin{abstract}
With the tremendous expansion of graphs data, node classification shows its great importance in many real-world applications. Existing graph neural network based methods mainly focus on classifying unlabeled nodes within fixed classes with abundant labeling. However, in many practical scenarios, graph evolves with emergence of new nodes and edges. Novel classes appear incrementally along with few labeling due to its newly emergence or lack of exploration. In this paper, we focus on this challenging but practical \emph{graph few-shot class-incremental learning} (GFSCIL) problem and propose a novel method called \textsc{Geometer}. Instead of replacing and retraining the fully connected neural network classifer, \textsc{Geometer} predicts the label of a node by finding the nearest class prototype. Prototype is a vector representing a class in the metric space. With the pop-up of novel classes, \textsc{Geometer} learns and adjusts the attention-based prototypes by observing the geometric proximity, uniformity and separability. Teacher-student knowledge distillation and biased sampling are further introduced to mitigate catastrophic forgetting and unbalanced labeling problem respectively. Experimental results on four public datasets demonstrate that \textsc{Geometer} achieves a substantial improvement of 9.46\% to 27.60\% over state-of-the-art methods.
\end{abstract}



\maketitle

\section{Introduction}
Graphs data are ubiquitously used to reveal the interactions among various entities, such as academic graphs, social networks, recommendation systems, etc. During the past several years, node classification~\cite{wang2020gcn,xhonneux2020continuous,DBLP:conf/kdd/WangWGG21,DBLP:conf/kdd/QiuCDZYDWT20,DBLP:conf/aaai/YouGYL21} has received considerable interests and achieved remarkable progress with the rise of graph neural networks (GNNs). In contrast, real-world networks evolve with the emergence of new nodes and edges, thereby generating novel classes. For example, in academic networks, the publication of new research papers produces new interdisciplines; Industrial development brings about new types of commodities in online e-commerce; The addition of new users leads to the emergence of new social groups. Classes of nodes are expanding incrementally and usually accompanied by few labeling due to its newly emergence or lack of exploration.



\begin{figure}
    \centering
    \includegraphics[width=\linewidth]{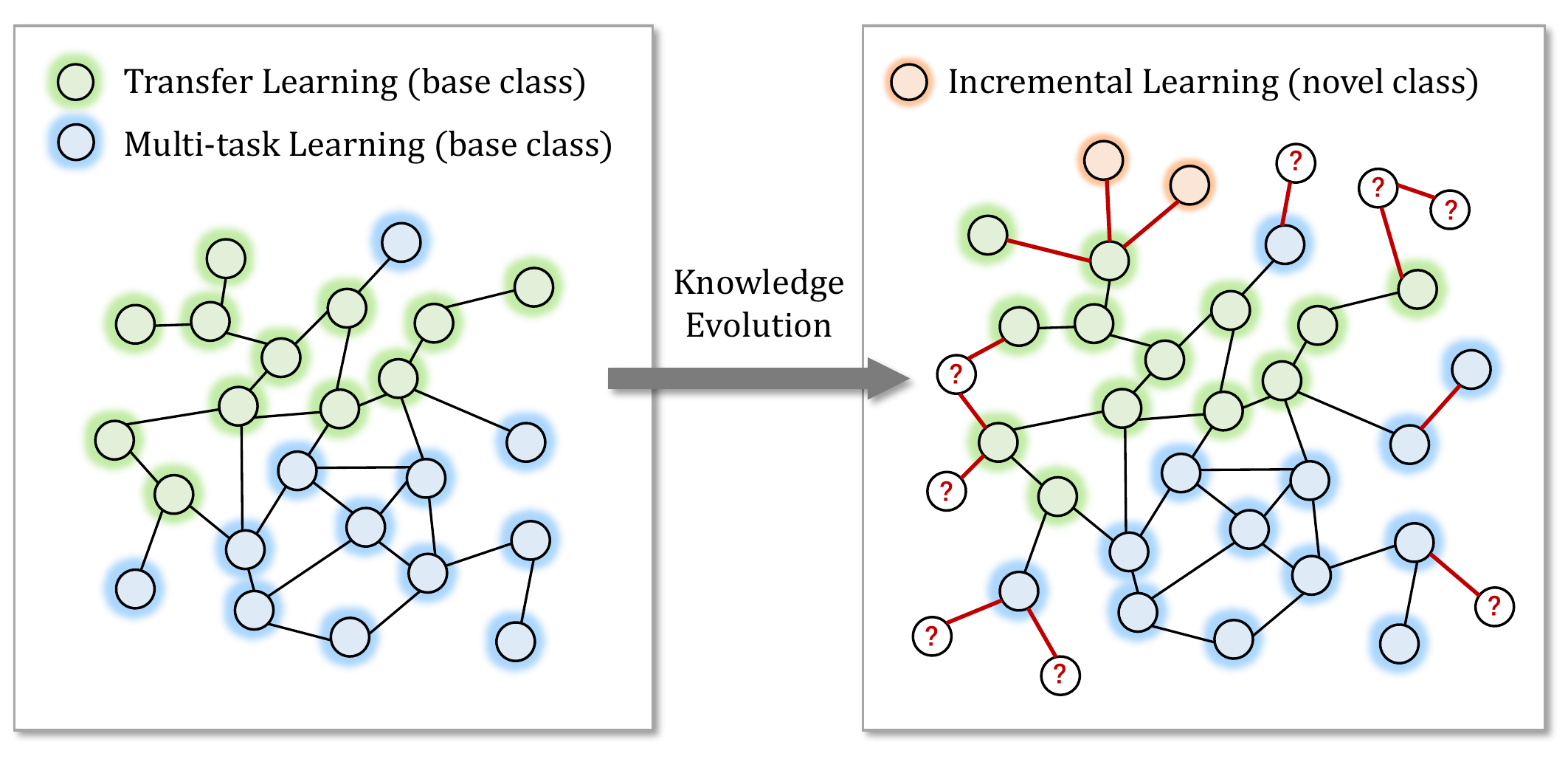}
    \caption{Illustration of GFSCIL problem on an academic graph. Nodes represent papers, edges represent citation relationships, and each paper belongs to a certain research field (node class).}
    \label{fig:toy_example}
\end{figure}

Take a toy academic graph in Figure \ref{fig:toy_example} for further illustration. Originally, there are abundant labeled nodes for ``Transfer Learning'' and ``Multi-task Learning'' (i.e., \emph{base classes}). With the knowledge evolution, new nodes appear and introduce additional citation relationships (edges). A new emerging research topic ``Incremental Learning'' (i.e., \emph{novel class}) has also turned up with few labeled nodes. A critical problem to be solved is to classify the remaining unlabeled nodes into either a \emph{base class} or a \emph{novel class}, where \emph{novel class} only have few labeled samples. We term this kind of node classification among all encountered classes (\emph{base classes} and \emph{novel classes} altogether) in dynamic graphs as \emph{graph few-shot class-incremental learning} (GFSCIL).

\textbf{Prior works.} Classical GNN-based methods~\cite{DBLP:conf/iclr/KipfW17,DBLP:conf/nips/Huang0RH18,DBLP:conf/aaai/BoWSS21} mainly focus on classifying the nodes within a set of fixed classes with abundant labelling. However, 
due to the \emph{few-shot} and \emph{class-incremental} nature, these methods fail to solve GFSCIL problem. 
Some advanced methods aim at addressing part of the problem of GFSCIL.
On one hand, to tackle the \emph{few-shot} node classification problems in graphs, MAML-based studies transfer the knowledge from \emph{base classes} to never-before-seen classes with only a handful of labeled information. 
Whereas, these methods~\cite{zhou2019meta,DBLP:conf/nips/HuangZ20,liu2021relative} all make a strong prior \emph{N-way K-shot} assumption that the unlabeled nodes belong to a fixed set of \emph{N novel classes}. 
Meanwhile, the classification of base classes and novel classes are separated into two models, 
which prevents from judging the results under a unified metric.
On the other hand, although \emph{class-incremental} learning has achieved significant progress in computer vision tasks~\cite{li2017learning,rebuffi2017icarl,hou2019learning}, class-incremental node classification in graphs has not been fully explored. Existing methods are dedicated to independent and identically distributed data (e.g., images), which has no explicit interations.
Graph data lies in non-Euclidean space and the network structure evolves dynamically.
The emergence of new edges changes and complicates the node correlations, thus bringing more challenges.






\textbf{Challenges.} A naive approach for GFSCIL is to finetune the base model on both \emph{base classes} and \emph{novel classes}. However, there are three main challenges that need to be addressed:
(1) \emph{How to find a way out of ``forgetting old''?} Catastrophic forgetting phenomenon~\cite{DBLP:journals/corr/GoodfellowMDCB13,kirkpatrick2017overcoming,DBLP:conf/icml/YapRB21} describes the performance degradation on old classes when incrementally learning novel classes. 
In GFSCIL, the growing number of novel classes makes the model suffer from forgetting base classes.
(2) \emph{How to overcome the unbalanced labeling between base classes and novel classes?} In GFSCIL, the labeling between large-scale base classes and few-shot novel classes is unbalanced. Directly training on few-shot samples may cause over-fitting problem. 
(3) \emph{How do we capture the dynamic structure as the network evolves?} The structure of graphs are highly dynamic in GFSCIL. The arrival of new nodes and edges make more complex connections, which is a big challenge for expressive node representations. 




\textbf{Our Work.} To address the aforementioned problems, we leverage the concept of metric learning and propose a new method for 
\underline{\textbf{G}}raph f\underline{\textbf{E}}w-Sh\underline{\textbf{O}}t Class-Incre\underline{\textbf{M}}ental L\underline{\textbf{E}}arning via Pro\underline{\textbf{T}}otyp\underline{\textbf{E}} \underline{\textbf{R}}epresentation, named \textsc{Geometer}. 
Instead of replacing and retraining the fully connected neural network classifer, \textsc{Geometer} predicts the ever-expanding class of a node by finding the nearest prototype representation. Prototype is a vector representing a class in the metric space. 
We propose class-level multi-head attention to learn the dynamic prototype representation of each class. When novel classes popping up, \textsc{Geometer} learns and adjusts the representation based on the geometric relationships of intra-class proximity, inter-class uniformity and inter-class separability in the metric space. 
In order to avoid \emph{forgetting old}, \textsc{Geometer} iteratively takes the previous model as the teacher, and guides the student model's representation of old classes with knowledge distillation. \textsc{Geometer} adopts pretrain-finetune paradigm with well-designed biased sampling strategy to further alleviate the impact of unbalanced labeling. 





To summarize, the main contributions of our works are as follows:
\begin{itemize}[left=1em]
    \item We investigate a novel problem for node classification: \emph{graph few-shot class-incremental learning} (GFSCIL). To the best of our knowledge, this is the first work to study this challenging yet practical problem.
    \item We propose a novel model \textsc{Geometer} to solve GFSCIL problem. With the novel classes popping up, \textsc{Geometer} learns and adjusts the attention-based prototypes based on the geometric relationships of proximity, uniformity and separability of representations.
    \item \textsc{Geometer} proposes teacher-student knowledge distillation and biased sampling strategy to further mitigate the catastrophic forgetting and unbalanced labeling in GFSCIL.
\end{itemize}
We conduct extensive experiments on four real-world node classification datasets to corroborate the effectiveness of our approach. \textsc{Geometer} achieves a substantial improvement of nearly 9.46\% to 27.60\% in multiple sessions of GFSCIL over state-of-the-art baselines.



\begin{figure*}[h]
    \centering
    \includegraphics[width=\linewidth]{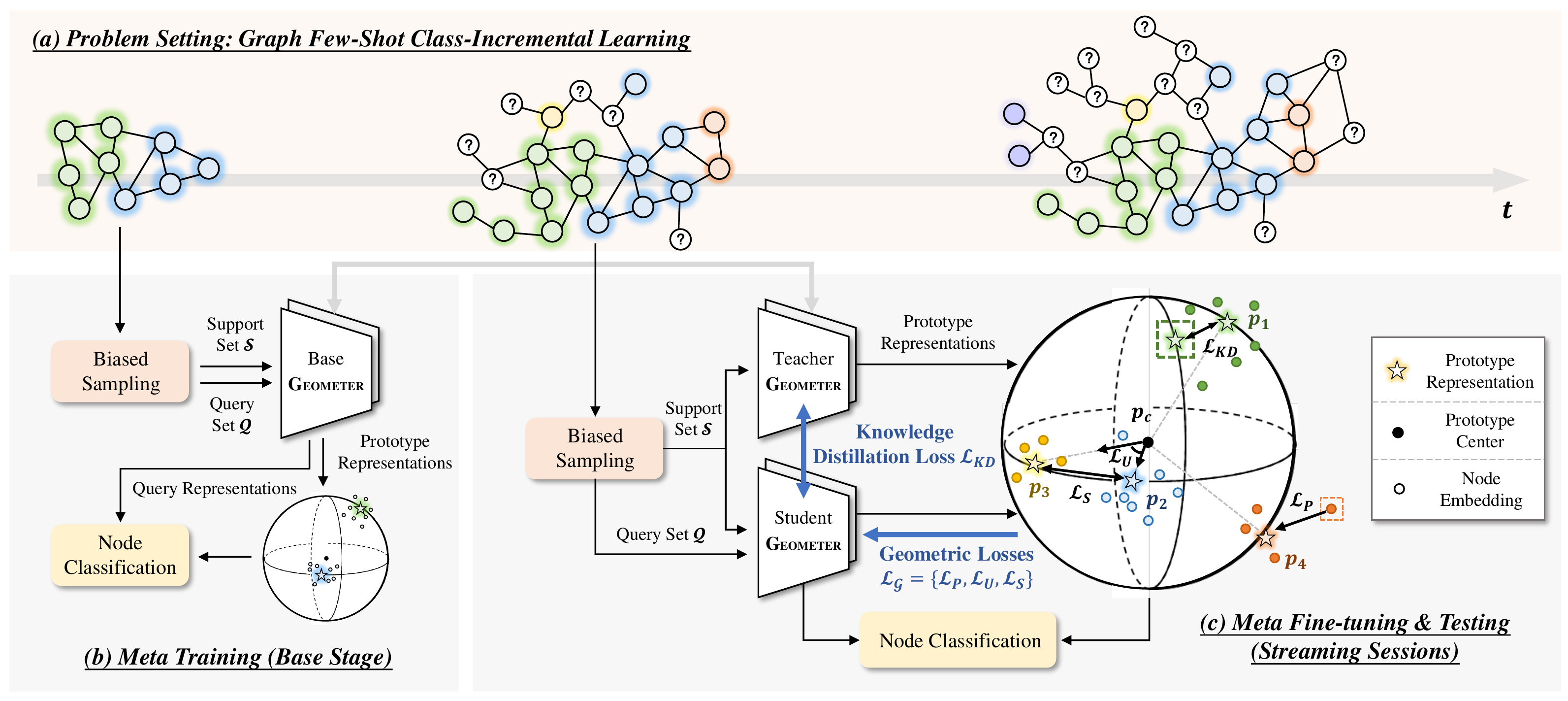}
    \caption{Overview of the proposed \textsc{Geometer} for Graph Few-Shot Class-Incremental Learning. (a) Problem setting of GFSCIL. With the arrival of nodes, the network structure has become more complex and novel node classes have been introduced (shown by different colors). (b) and (c) show the episode meta learning process with biased sampling strategy at base stage and streaming sessions. Two different loss functions $\mathcal{L}_{\mathcal{G}}$ and $\mathcal{L}_{KD}$ are utilized for the update of the metric space.}
    \label{fig:system_model}
\end{figure*}

\section{Related Work}
In this section, we briefly introduce the relevant research lines of our work, namely few-shot node classification and class-incremental learning.

\subsection{Few-Shot Node Classification}

In recent years, few-shot node classification on graph has attracted increasing attention. These works can be categorized into two types: (1) optimization based approaches, and (2) metric based approaches. Optimization-based approaches leverage MAML~\cite{finn2017model} to learn a better GNN initialization on base classes, and quickly adapt to novel classes with few-shot samples. Meta-GNN~\cite{zhou2019meta} firstly incorporates the meta-learning paradigm into GNNs for few-shot node classification. G-Meta~\cite{DBLP:conf/nips/HuangZ20} proposes to use local subgraphs to learn the transferable knowledge across tasks. Liu et al.~\cite{liu2021relative} 
further design the relative and absolute embedding of nodes and achieves promising performance.
However, these method divide the classification of novel classes and old classes into two seperate models, which cannot carry out a unified classification of the unknown nodes in GFSCIL.
Metric-based methods propose to learn a transferable metric space, which is closely related to our work. Ding et al. propose GPN~\cite{ding2020graph} for few-shot learning on attributed graphs by combining prototype network with GNNs. Yao et al.~\cite{yao2020graph} incorporate prior knowledge learned from auxiliary graphs to further transfer the knowledge to a new target graph.
Whereas, the growing of new labels in GFSCIL will make the prototypes overlapping, and the accuracy of the classification will decline sharply.




\subsection{Class-Incremental Learning}

Class-incremental learning aims to learn a unified classifier 
to recognize the ever-expanding classes over time, which is extensively studied in the field of computer vision~\cite{li2017learning,rebuffi2017icarl,castro2018end,hou2019learning}. 
iCaRL~\cite{rebuffi2017icarl} adopts an ``episodic memory'' of class exemplars and incrementally learns the nearest-neighbor classifier for novel classes. 
Castro et al.~\cite{castro2018end} propose a distillation measure to retain the knowledge of old classes, and combines it with the cross-entropy loss for end-to-end training. 
Hou et al.~\cite{hou2019learning} propose the multi-class incremental setting and raises the imbalance challenge between old classes and novel classes.
However, among these works, the training samples of novel classes are all large-scale. In many real-world scenarios, the novel classes often lack of labeling due to its newly emergence or lack of exploration. Recently, the FSCIL problem has just been put forward in image classification~\cite{tao2020few,cheraghian2021semantic}. Tao et al.~\cite{tao2020few} firstly propose this FSCIL problem and utilize a neural gas (NG) network to learn and maintain the topology of the feature manifold of various classes. Cheraghian et al.~\cite{cheraghian2021semantic} further introduce a distillation algorithm with semantic information.
Except in the field of computer vision, few-shot class-incremental learning also shows practical significance in graphs and remains an under-explored problem. To the best of our knowledge, this is the first study of FSCIL for node classification in graphs, which we denoted as GFSCIL.


 
\section{Problem Statement}

In this section, we provide problem statement and definitions. In the base stage, we have an initial graph $\mathcal{G}^{base}$. In streaming sessions, suppose we have $T$ snapshots of evolving graph, denoting as $\mathcal{G}^{stream} = \{\mathcal{G}^1, \cdots, \mathcal{G}^T\}$. Take the $t$-th session as an example, its corresponding graph represents as $\mathcal{G}^t = (\mathcal{V}^t, \mathcal{E}^t, \mathbf{X}^t)$. Suppose we have $N_t$ nodes and $M_t$ edges. $\mathcal{V}^t$ is the node set $\{v_1, v_2, \cdots, v_{N_t}\}$, and $\mathcal{E}^t$ is the edge set $\{e_1, e_2, \cdots, e_{M_t}\}$. 
The feature vector of node $v_i$ is represented as $\bm{x}_{i} \in \mathbb{R}^{d}$, and $\mathbf{X}^t = \{\bm{x}_1, \cdots, \bm{x}_{N_t}\} \in \mathbb{R}^{{N_t}\times d}$ denotes all the node features. We denote $\{\mathcal{C}^{base}, \mathcal{C}^{1}, \cdots, \mathcal{C}^{T}\}$ as sets of classes from base stage to the $T$-th streaming session. $\mathcal{C}^{base}$ is the set of base classes with large training samples. In $t$-th streaming session, $\Delta \mathcal{C}^t$ novel classes are introduced with few-shot samples, where $\forall i,j, \Delta \mathcal{C}^{i} \cap \Delta \mathcal{C}^{j} = \varnothing$ and $\mathcal{C}^{i} = \mathcal{C}^{i-1} + \Delta \mathcal{C}^{i}$.
We denote the totally encountered class in $t$-th session as $\mathcal{C}^{t} = \mathcal{C}^{base} + \sum_{i=1}^{t} \Delta \mathcal{C}^i$.

\begin{myProb}
\textbf{Graph Few-Shot Class-Incremental Learning} In $t$-th streaming session, we denote $\Delta \mathcal{C}^t$ novel classes with $K$ labeled nodes as the $\Delta \mathcal{C}^t$-way $K$-shot GFSCIL problem. 
The labeled training samples are denoted as support sets $\mathcal{S}$. 
Another batch of nodes to predict their corresponding label are denoted as query sets $\mathcal{Q}$.
After training on the support sets $\mathcal{S}$ of $t$-th session, the GFSCIL problem is tested to classify unlabel nodes of query sets $\mathcal{Q}$ into all encountered classes $\mathcal{C}^{t}$.
\end{myProb}

\begin{myDef}
\textbf{Prototype Representation} A prototype representation is a representative embedding of one class. The node embeddings of one class tend to cluster around its prototype representation in the same metric space. Prototype representation is first proposed in {\rm{~\cite{snell2017prototypical}}}, which regards the mean of its support set as class’s prototypes.
\end{myDef}

\section{Methodology}

We first give an overview of the proposed \textsc{Geometer}, as illustrated in Figure \ref{fig:system_model}.
\textsc{Geometer} intends to predict the node class by finding the nearest \emph{attention-based prototype representation}. When novel classes emerging, we learn and adjust prototypes based on \emph{geometric metric learning} and \emph{teacher-student knowledge distillation}.
Our approach follows the \emph{episode meta learning} process, and different biased sampling are designed to overcome the unbalanced labeling among base classes and novel classes.





\subsection{Attention-based Prototype Representation}

\begin{figure}
    \centering
    \includegraphics[width=0.8\linewidth]{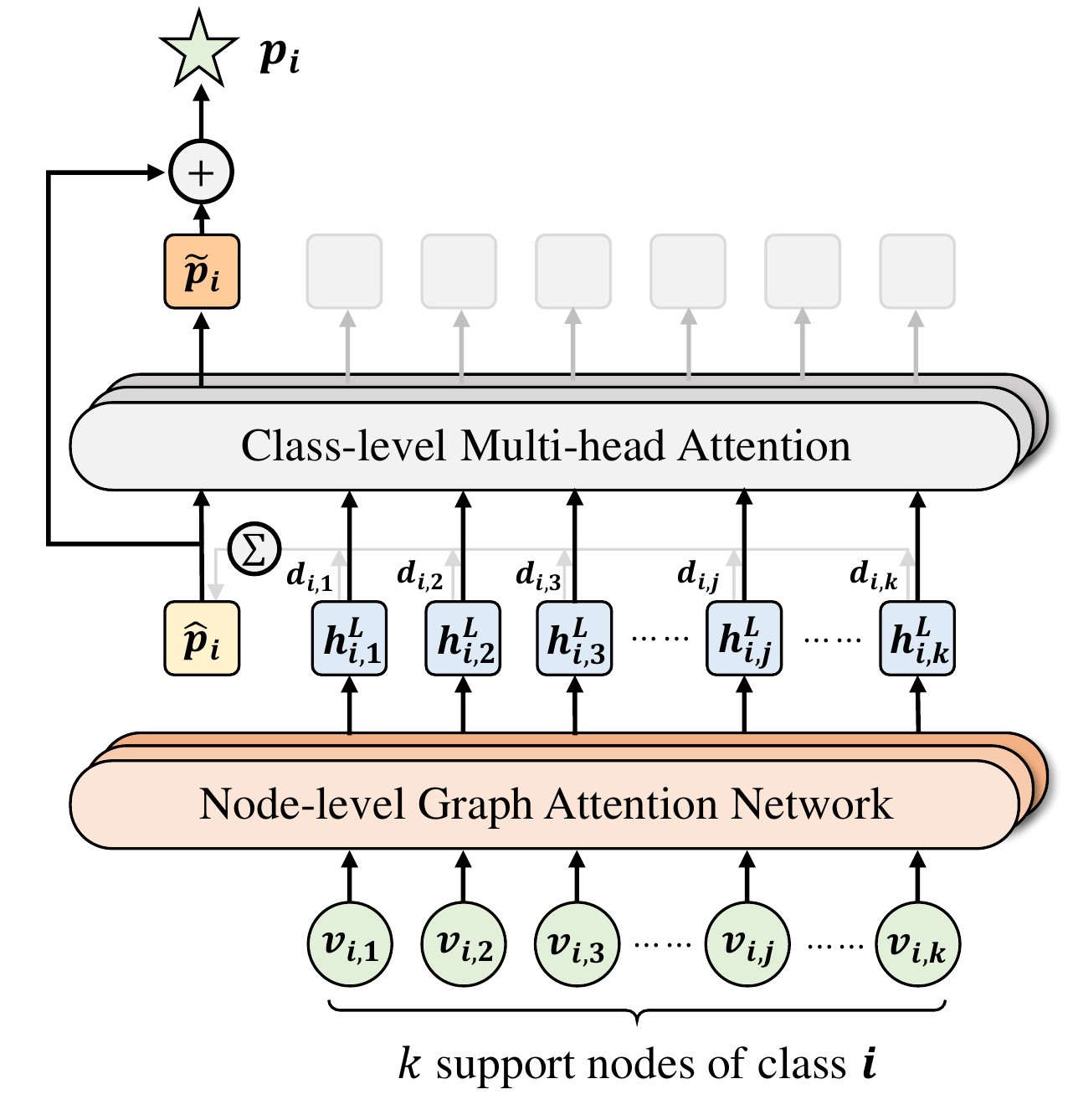}
    \caption{Attention-based Prototype Representation Model}
    \label{fig:proto_repre}
\end{figure}

The evolution of the network makes the influence of nodes unequal and non-static.
In addition, weakly-labeled few-shot data usually contains a significant amount of noise.
Therefore, direct average of support node features cannot be fully representative and is highly vulnerable to the noise or outliers. In order to learn the expressive prototype representation of each class, we propose a two-level attention-based prototype representation learning method as shown in Figure \ref{fig:proto_repre}. 



\subsubsection{Node-level Graph Attention Network} Graph neural network is typically expressed as a message-passing process in which information can be passed from one node to another along edges directly. \emph{Node-level graph attention network} $f_{\mathcal{G}}(\cdot)$ computes a learned edge weight by performing masked attention mechanism~\cite{DBLP:conf/iclr/VelickovicCCRLB18}. The attention score $\alpha_{i j}$ between node $v_i$ and $v_j$ is normalized across all node $v_i$'s neighbors $\mathcal{N}_i$ as 
\begin{equation}
    \alpha_{i j}=\frac{\exp \left(\operatorname{LeakyReLU}\left({\mathbf{a}}^{T}\left[\mathbf{W} \bm{h}^{l}_{i} \| \mathbf{W} \bm{h}^{l}_{j}\right]\right)\right)}{\sum_{k \in \mathcal{N}_{i}} \exp \left(\operatorname{LeakyReLU}\left(\mathbf{a}^{T}\left[\mathbf{W} \bm{h}^{l}_{i} \| \mathbf{W} \bm{h}^{l}_{k}\right]\right)\right)},
\end{equation}
where $\bm{h}^{l}_{i}$ and $\bm{h}^{l}_{j}$ represent the node features of $l$-th GNN layer,
$\mathbf{a} \in \mathbb{R}^{2d^{\prime}}$ and $\mathbf{W} \in \mathbb{R}^{d^{\prime} \times d}$ are weight matrices. $\|$ denotes vector concatenation.
Then, graph attention network computes a weighted average of the transformed features of the neighbor nodes as the new representation, followed by a nonlinear function $\sigma$. The $(l+1)$-th layer hidden state of node $v_{i}$ is calculated via
\begin{equation}
    \bm{h}_{i}^{l+1}=\sigma(\sum_{j \in \mathcal{N}_{i}} \alpha_{i j} \cdot \mathbf{W} \bm{h}_{j}^{l}).
\end{equation}

We denote the $L$-th layer hidden state output of node $v_i$ as $f_{\mathcal{G}}(\bm{x}_i) = \bm{h}_{i}^{L}$.
As usual, we build a 2-layer graph attention network for feature extraction.






\subsubsection{Class-level Multi-head Attention}
Due to the dynamic structure and stochastic label noise, \textsc{Geometer} proposes \emph{class-level multi-head attention} to learn the class prototype as follows. In streaming fashion of GFSCIL, 
the importance of nodes changes dynamically as the network evolves.
The degree centrality is one of simplest way to measure the importance of nodes. 
The large-degree nodes are often referred to as hubs, which has a stronger influence in the networks.
Therefore, an initial prototype $\bm{\hat{p}}_{i}$ of class $i$ is calculated by degree-based weighted-sum of support node embeddings:

\begin{equation}
    \bm{\hat{p}}_{i} = \sum_{j \in \mathcal{S}_i} \frac{\text{degree}(v_j)}{\sum_{j^{\prime} \in \mathcal{S}_i} \text{degree}(v_{j^{\prime}})} \cdot f_{\mathcal{G}}(\bm{x}_{j}),
\end{equation}
where $\mathcal{S}_{i}$ is the support set of class $i$, $f_\mathcal{G}(\bm{x}_j)$ is the node representation of $v_j$ obtained by \emph{node-level graph attention network}, $\text{degree}(v_j)$ is the degree centrality of node $v_j$. Apart from considering the structural information, i.e. degree centrality, different support node features plays an important role in learning a representative class prototype.
In order to fully characterize the relationship between node features and prototypes, \emph{class-level multi-head attention} calculates the attention score between the initial prototype and support node representations to obtain an expressive prototype representation. To be specific, we take the linear transformation of initial prototype $\bm{\hat{p}}_{i}$ as the query $\textbf{Q}$ and then concatenate the initial prototype and support node representations as $\bm{h}_{i}^{spt}$:
\begin{equation}
    \bm{h}_{i}^{spt} = \textsc{Concatenate}(\bm{\hat{p}}_i, \|_{j\in \mathcal{N}_i} f_{\mathcal{G}}(\bm{x}_j)).
\end{equation}
We take the linear transformation of $\bm{h}_{i}^{spt}$  as the key $\textbf{K}$ and value $\textbf{V}$.
\textsc{Geometer} adopts the scaled dot-product attention~\cite{vaswani2017attention}, which is calculated via
\begin{equation}
    \textsc{Attention}(\bm{Q},\bm{K},\bm{V}) = \text{softmax}(\frac{\bm{Q}\bm{K}^{T}}{\sqrt{d_k}})\bm{V}, 
\end{equation}
where $\bm{Q} = \mathbf{W}_Q \bm{\hat{p}}_{i}$, $\bm{K} = \mathbf{W}_K \bm{h}_{i}^{spt}$, $\bm{V} = \mathbf{W}_V \bm{h}_{i}^{spt}$, $\mathbf{W}_Q, \mathbf{W}_K, \mathbf{W}_V$ are three weight matrices, and $d_k$ is the dimension of $\bm{Q}$ and $\bm{K}$.
Finally, the residual connection is adopted to obtain the final prototype representation $\bm{p}_i$ of class $i$: 
\begin{equation}
    \bm{p}_i = \bm{\hat{p}}_i + \textsc{Attention}(\bm{\hat{p}}_i, \bm{h}_{i}^{spt}, \bm{h}_{i}^{spt}).
\end{equation}





\subsection{Geometric Metric Learning}

In GFSCIL problem, as the graph evolves, novel node classes obtain new prototypes in the metric space. With the increase of growing classes, the performance of node classification is greatly reduced due to the overlapping of prototype representations.
As a consequence of few-shot samples of novel classes, parameter update only based on node classification results is prone to overfitting novel classes, 
or is greatly affected by the support set sample distribution. 
Therefore, we propose to learn the prototype representation from geometric relationships. As shown in Figure \ref{fig:system_model}(c), we propose geometric loss functions from three aspects: intra-class proximity, inter-class uniformity and inter-class separability. 

\subsubsection{Intra-Class Proximity}
Intra-class proximity indicates that the nodes of same classes should be closely clustered.
Therefore, in the metric space, the distance between the node embedding and its corresponding class prototype representation should be relatively close.
We use squared Euclidean distance $d(\cdot)$ to measure the distance between node features and class prototype, and define the intra-class proximity loss $\mathcal{L}_{P}$ as follows:
\begin{equation}
    \mathcal{L}_{P}=\sum_{k=1}^{\| \mathcal{C}^{k}\|}  \frac{\alpha_{k}}{n_{k}} \sum_{i=1}^{n_{k}}-\log \frac{\exp \left(-d\left(f_{\mathcal{G}}\left(\bm{x}_{i}\right), \bm{p}_{k}\right)\right)}{\sum_{k^{\prime}\in \mathcal{C}^{k}} \exp \left(-d\left(f_{\mathcal{G}}\left(\bm{x}_{i}\right), \bm{p}_{k^{\prime}}\right)\right)},
\end{equation}
where $\| \mathcal{C}^{k}\|$ is the total number of encountered classes up to $k$-th streaming session, $n_k$ is the number of node samples of class $k$. $\alpha_k \in \left[0,1\right]$ is a weighting factor, which is used to adjust the impact of unbalanced labeling of base classes and novel classes in total loss function.
    
\subsubsection{Inter-Class Uniformity}

Inter-class uniformity describes the positional uniformity of different prototypes in metric space. Specifically, a prototype center $p_c$ is denoted by the mean of all prototypes:
\begin{equation}
    \bm{p}_{c}=\frac{1}{\left\|\mathcal{C}^{k}\right\|} \sum_{i=1}^{\left\|\mathcal{C}^{k}\right\|} \bm{p}_{i},
\end{equation}
Geometrically, taking the prototype center $p_c$ as the coordinate origin, each normalized prototype relative to the center $\frac{\bm{p}_{j}-\bm{p}_{c}}{\left\|\bm{p}_{j}-\bm{p}_{c}\right\|}, \forall j$
is distributed on a unit sphere.
As the prototypes of the novel classes are increasingly projected onto the unit sphere, we propose that the distribution of prototypes should tend to be uniform. 
The distribution of class prototypes are adjusted based on the division of sphere angle. 
Therefore, we define the following inter-class uniformity loss function $\mathcal{L}_{U}$ based on cosine similarity distance as
\begin{equation}
    \mathcal{L}_{U}=\frac{1}{\left\|\mathcal{C}^{k}\right\|} \sum_{i=1}^{\left\|\mathcal{C}^{k}\right\|} \left\{ 1+\max _{j \in \{\mathcal{C}^k\} \backslash i}\left[\frac{\left(\bm{p}_{i}-\bm{p}_{c}\right)}{\left\|\bm{p}_{i}-\bm{p}_{c}\right\|} \cdot \frac{(\bm{p}_{j}-\bm{p}_{c})}{\|\bm{p}_{j}-\bm{p}_{c}\|}\right] \right\},
\end{equation}
where 1 is used as a bias to ensure that the value is always non-negative, and the purpose of taking the maximum value of cosine similarity here is to focus on the angular distribution of adjacent prototypes. By defining the inter-class uniformity, the unbalanced labeling class prototypes are inclined to be evenly distributed on the sphere. Especially for novel classes with few-shot labeled nodes, the inter-class geometric relations provide important guidance for the learning of prototype representations.

\subsubsection{Inter-Class Separability}
Before finetuning, the feature extractor is more suitable for old classes representation. The prototypes of novel classes are likely to overlap with the old class prototypes, which greatly affects the accuracy of node classification. 
Therefore, we propose inter-class separability in geometric metric learning, which describes that the prototypes of novel classes and old classes should keep a distance in the metric space. The inter-class separability loss $\mathcal{L}_{S}$ is denoted as
\begin{equation}
    \mathcal{L}_{S}=\frac{1}{\Delta \mathcal{C}^{k}} \sum_{i \in \Delta \mathcal{C}^{k}} \min _{j \in \mathcal{C}^{k-1}} \exp \left(-d\left(\bm{p}_{i}, \bm{p}_{j}\right)\right),
\end{equation}
where $d(\cdot)$ is the squared euclidean distance, $\mathcal{C}^{k-1}$ is the set of labels of $(k-1)$-th session, and $\Delta \mathcal{C}^{k}$ is the novel classes of $k$-th
session. The definition of inter-class separability is a supplement to inter-class uniformity. With the addition of novel classes, in order to avoid the error diffusion caused by the inaccurate classification of old categories, expanding the distance between old and novel class prototypes in metric space helps to further enhance the classification accuracy in GFSCIL settings.

\subsection{Teacher-Student Knowledge Distillation}
Due to the class-incremental nature of GFSCIL problem, \textsc{Geometer} applies the idea of teacher-student knowledge distillation to further mitigate ``forgetting old'' during finetuning.
In contrast to the classic teacher-student knowledge distillation techniques~\cite{hinton2015distilling,10.1145/3447548.3467319,10.1145/3394486.3403234} used to compress large model into lightweight model with better inference efficiency, we regard the model before streaming as the teacher model and new model as the student model. 
Knowledge distillation technique is used to transfer the classification ability of previous model, while preserving the interrelationships of the old classes in the metric space.
Temperature-scaled softmax~\cite{li2017learning} is utilized to soften the old classes logits of teacher model and student model. The modified logits $y^{\prime(i)}$ of class $i$ by applying a temperature scaling function in the softmax are calculated as
\begin{equation}
    y^{\prime(i)} = \frac{\exp \left(d\left(f_{\mathcal{G}}\left(\bm{x}^{(i)}\right), \bm{p}_{i}\right) / \tau\right)}{\sum_{j} \exp \left(d\left(f_{\mathcal{G}}\left(x^{(i)}\right), \bm{p}_{j}\right) / \tau\right)},
\end{equation}
where $\tau$ is the temperature factor. Generally, we set $\tau > 1$ to increase the weight of smaller logit values and encourages the network to better reveal inter-class relationships learned by the teacher model. \textsc{Geometer} proposes to calculate the KL-divergence of the softened logits to make the student model gain the experience of classifying old classes $\mathcal{C}^{t-1}$ from teacher model. The 
teacher-student knowledge distillation loss $\mathcal{L}_{K D}$ on $k$-th session is calculated as
\begin{equation}
    \mathcal{L}_{K D}=\frac{1}{\left\|\mathcal{C}^{k-1}\right\|} \sum_{i=1}^{\left\| \mathcal{C}^{k-1}\right\|} y_{S}^{\prime(i)} \cdot \log \left(\frac{y_{S}^{\prime(i)}}{y_{T}^{\prime(i)}}\right),
\end{equation}
where $\mathcal{C}^{k-1}$ is the set of old classes of $(k-1)$-th streaming session and $y_{S}^{\prime(i)}$ and $y_{T}^{\prime(i)}$ are the modified logits. \textsc{Geometer} proposes the teacher-student knowledge distillation to avoid the catastrophic forgetting caused by the growing addition of novel classes. Meanwhile, it makes \textsc{Geometer} a system of checks and balances related to geometric losses.




\subsection{Episode Meta Learning}

In this section, we discuss the learning process of \textsc{Geometer}. We adopt the episode paradigm in learning process, which has shown great promise in few-shot learning. Instead of directly training or finetuning on batches of data, a set of
tasks $\mathcal{T}$ are generated by imitating the \emph{N-way-K-shot} few-shot scenario.
Each task $\mathcal{T}_i \in \mathcal{T}$ includes support set $\mathcal{S}_i$ and query set $\mathcal{Q}_i$.
In GFSCIL setting, two different \emph{bias sampling strategies} are adopted in both pretraining and finetuning stages. 

In the base stage, all base classes have a large number of training samples. However, in streaming sessions, the novel classes follows few-shot labeling. \textsc{Geometer} adopts biased sampling strategy in pretraining stage 
to generate class-imbalanced support sets by mimicking the circumstances encountered during finetuning. Specifically, the number of sampling number of each class in support set $\mathcal{S}$ follow a uniform distribution $\textbf{U}[1, K_{max}]$. The size of query set still maintains a fixed number $K_{qry}$. 
During meta training, intra-class proximity and inter-class uniformity loss are utilized. Therefore, the loss function $\mathcal{L}_{\text{train}}$ is as follows, with hyper-parameters $\lambda_{P}$ and $\lambda_{U}$:
\begin{equation}
    \mathcal{L}_{\text{train}} = \lambda_{P}\mathcal{L}_{P} + \lambda_{U}\mathcal{L}_{U}.
\end{equation}

In the streaming sessions, the biased sampling adopts different strategies to obtain the class-imbalanced query set $\mathcal{Q}$. In order to avoid ``forgetting old" and ``overfitting new" problem, a higher proportion of the old samples will be sampled when sampling the query set. 
This helps to fully retain the classification accuracy on old classes during meta fine-tuning. During meta-finetuning, three geometric losses and the knowledge distillation losses are taken into account, and the loss function $\mathcal{L}_{\text{finetune}}$ is calculated as
\begin{equation}
    \mathcal{L}_{\text{finetune}} = \lambda_{P}\mathcal{L}_{P} + \lambda_{U}\mathcal{L}_{U} + \lambda_{S}\mathcal{L}_{S} + \lambda_{KD}\mathcal{L}_{KD},
\end{equation}
where $\lambda_{P}$, $\lambda_{U}$, $\lambda_{S}$ and $\lambda_{KD}$ are hyper-parameters.

\section{Experiment}

\subsection{Experimental Setup}

\subsubsection{Datasets}
We evaluate the proposed \textsc{Geometer} on four real-world representative datasets: Cora-ML, Flickr, Amazon and Cora-Full. We summarize the statistics of these datasets in Table \ref{tab:dataset}.
A detailed description of these four datasets is provided in Appendix \ref{appendix-dataset}.

\begin{table}[h]
\centering
\caption{Statistics of datasets used in the experiments}
\label{tab:dataset}
\resizebox{\linewidth}{!}{%
\begin{tabular}{cccccc}
\toprule
\textbf{Dataset} & \textbf{Data Field} & \textbf{Nodes} & \textbf{Edges} & \textbf{Features} & \textbf{Class} \\ \midrule
Cora-ML & Academic       & 2,995  & 16,316  & 2,879  & 7  \\
Flickr  & Social network & 7,575  & 479,476 & 12,047 & 9  \\
Amazon  & E-commerce     & 13,752 & 491,722 & 767    & 10 \\
Cora-Full    & Academic       & 19,793 & 126,842 & 8,710  & 70 \\ \bottomrule
\end{tabular}%
}
\end{table}

\subsubsection{Experiment Settings}

We divide the dataset into base stage and several streaming sessions respectively. 
For Cora-ML, Flickr and Amazon datasets, we select five classes as novel classes and the rest as base classes, and adopt the \emph{1-way 5-shot} GFSCIL setting, which means we have 6 sessions (i.e., 1 base + 5 novel) in total. While for Cora-Full dataset, we adopt \emph{5-way 5-shot} GFSCIL setting, by choosing 20 classes as base classes and splitting the remaining 50 classes into 10 streaming sessions. 
We set 2-layer GNNs with 512 neurons of hidden layer. The learning rate of base class is 1e-3, and the learning rate during fine-tuning is 1e-4. The temperature factor $\tau$ is 2.



\subsubsection{Baseline Methods}

We compare the proposed method with following baselines:

\begin{itemize}[left=1em]
    \item \textbf{GAT (FT)}: Graph attention network (GAT)~\cite{DBLP:conf/iclr/VelickovicCCRLB18} is one of the state-of-the-art methods for node classification. We first pretrain a 2-layer GAT model with a fully connected neural network classifer on the base classes. During streaming sessions, we replace and retrain the parameters of the fully connected neural network classifer on the support set of both base classes and novel classes.
    \item \textbf{GAT+ (FT)}: It adopts the same architecture of GAT (FT). The difference is that we fine-tune all training parameters on support set on different streaming sessions.
    \item \textbf{GPN}~\cite{ding2020graph}: GPN is a superior method for few-shot node classification. It exploits graph neural networks and meta-learning on attributed networks for metric-based few-shot learning.
    \item \textbf{GFL}~\cite{yao2020graph}: It is the first work that resorts to knowledge transfer to improve semi-supervised node classification in graphs. It integrates local node-level and global graph-level knowledge to learn a transferable metric space, which is shared between auxiliary graphs and the target. 
    \item \textbf{PN*}~\cite{snell2017prototypical}: Prototype Network firstly proposed for few-shot image classification. We adopt the key idea and implement \textbf{PN*} for node classification.
    \item \textbf{PN* (FT)}: The training process is the same as PN*, but in the test process it fine-tunes all trained parameters before making prediction on the query set.
    \item \textbf{iCaRL*}~\cite{rebuffi2017icarl}: iCaRL is a class-incremental methods for image classification. We replace the feature extractor as a two-layer GAT network.
\end{itemize}

We only modify the dataset partition to satisfy the graph few-shot class-incremental settings for GAT, GPN and GFL, while other settings are the same as its original implementation. PN and iCaRL are two methods proposed in image classification tasks. To explore its performance on node classification, we utilize GNN-based backbone for feature extraction, and we marker with asterisk(*) for clarification.  
The above baselines can be can be summarized into four categories: (1) GNN methods with fully connected neural network classifer, which is a widely used architecture for node classification. (2) Representative graph few-shot learning models includes GPN and GFL. (3) Prototype network methods. (4) Class-incremental learning baselines.

\begin{table*}[]
\centering
\caption{Comparison results of node classification accuracy in GFSCIL settings on Cora-Full and Flickr dataset. \textsc{Geometer}’s improvement is calculated relative to the best baseline.}
\label{tab:cora_flickr}
\resizebox{\linewidth}{!}{%
\begin{tabular}{@{}cccccccccc@{}}
\toprule
 &
  \multicolumn{8}{c}{\textbf{Cora-Full (5-way 5-shot GFSCIL setting)}} &
   \\ \cmidrule(lr){2-9}
\multirow{-2}{*}{\textbf{Session}} &
  \textbf{GAT (FT)} &
  \textbf{GAT+ (FT)} &
  \textbf{GPN} &
  \textbf{GFL} &
  \textbf{PN*} &
  \textbf{PN* (FT)} &
  \textbf{iCaRL*} &
  \multicolumn{1}{l}{\cellcolor[HTML]{EFEFEF}\textbf{\textsc{Geometer} (Ours)}} &
  \multirow{-2}{*}{\textbf{impr.}} \\ \midrule
Base &
  \cellcolor[HTML]{EFEFEF}\textbf{80.53±1.32\%} &
  81.11±0.79\% &
  73.82±1.94\% &
  76.02±0.94\% &
  74.88±0.89\% &
  74.18±0.72\% &
  73.92±1.06\% &
  79.88±0.96\% &
  -1.52\% \\ \midrule
Session 1 &
  33.13±2.51\% &
  37.10±1.51\% &
  55.95±1.52\% &
  60.50±0.74\% &
  56.60±1.11\% &
  58.07±0.92\% &
  59.33±1.79\% &
  \cellcolor[HTML]{EFEFEF}\textbf{69.48±1.66\%} &
  +14.84\% \\
Session 2 &
  25.39±1.59\% &
  26.34±0.98\% &
  49.49±1.57\% &
  52.85±1.88\% &
  48.59±0.66\% &
  53.97±0.97\% &
  54.05±0.70\% &
  \cellcolor[HTML]{EFEFEF}\textbf{61.34±0.92\%} &
  +13.49\% \\
Session 3 &
  17.48±1.59\% &
  17.41±1.43\% &
  43.41±1.66\% &
  43.88±2.84\% &
  39.70±1.25\% &
  43.76±0.92\% &
  44.65±0.55\% &
  \cellcolor[HTML]{EFEFEF}\textbf{53.61±0.81\%} &
  +20.07\% \\
Session 4 &
  12.09±1.35\% &
  12.12±0.78\% &
  39.03±1.29\% &
  38.22±1.81\% &
  37.33±2.07\% &
  41.83±0.91\% &
  40.52±1.56\% &
  \cellcolor[HTML]{EFEFEF}\textbf{48.24±1.46\%} &
  +15.30\% \\
Session 5 &
  10.04±1.56\% &
  8.54±0.39\% &
  35.12±1.98\% &
  38.69±2.50\% &
  32.66±2.01\% &
  37.35±0.73\% &
  36.25±1.06\% &
  \cellcolor[HTML]{EFEFEF}\textbf{44.97±1.03\%} &
  +16.23\% \\
Session 6 &
  8.63±0.88\% &
  7.01±0.80\% &
  33.34±1.35\% &
  33.94±3.53\% &
  30.83±2.09\% &
  36.56±0.84\% &
  33.46±1.16\% &
  \cellcolor[HTML]{EFEFEF}\textbf{42.93±0.88\%} &
  +17.42\% \\
Session 7 &
  7.76±0.62\% &
  5.79±0.42\% &
  31.98±1.03\% &
  32.60±1.65\% &
  29.52±1.92\% &
  34.70±0.20\% &
  32.68±1.44\% &
  \cellcolor[HTML]{EFEFEF}\textbf{42.82±1.14\%} &
  +23.40\% \\
Session 8 &
  6.99±0.72\% &
  5.38±0.49\% &
  30.63±1.64\% &
  28.32±1.78\% &
  28.39±1.97\% &
  33.97±1.24\% &
  31.02±1.48\% &
  \cellcolor[HTML]{EFEFEF}\textbf{41.01±0.96\%} &
  +20.72\% \\
Session 9 &
  5.95±0.75\% &
  4.49±0.40\% &
  30.53±1.80\% &
  21.95±1.71\% &
  27.65±2.19\% &
  33.71±0.75\% &
  30.37±1.76\% &
  \cellcolor[HTML]{EFEFEF}\textbf{40.49±0.97\%} &
  +20.11\% \\
Session 10 &
  5.51±0.95\% &
  3.92±0.61\% &
  28.33±1.48\% &
  21.77±1.50\% &
  26.07±1.89\% &
  31.67±0.55\% &
  29.21±1.71\% &
  \cellcolor[HTML]{EFEFEF}\textbf{39.32±0.78\%} &
  +24.15\% \\ \midrule
\multicolumn{1}{l}{} &
  \multicolumn{1}{l}{} &
  \multicolumn{1}{l}{} &
  \multicolumn{1}{l}{} &
  \multicolumn{1}{l}{} &
  \multicolumn{1}{l}{} &
  \multicolumn{1}{l}{} &
   &
  \multicolumn{1}{l}{} &
  \multicolumn{1}{l}{} \\ \midrule
 &
  \multicolumn{8}{c}{\textbf{Flickr (1-way 5-shot GFSCIL setting)}} &
   \\ \cmidrule(lr){2-9}
\multirow{-2}{*}{\textbf{Session}} &
  \textbf{GAT (FT)} &
  \textbf{GAT+ (FT)} &
  \textbf{GPN} &
  \textbf{GFL} &
  \textbf{PN*} &
  \textbf{PN* (FT)} &
  \textbf{iCaRL*} &
  \multicolumn{1}{l}{\cellcolor[HTML]{EFEFEF}\textbf{\textsc{Geometer} (Ours)}} &
  \multirow{-2}{*}{\textbf{impr.}} \\ \midrule
Base &
  62.16±1.30\% &
  61.41±1.55\% &
  72.80±1.13\% &
  \cellcolor[HTML]{EFEFEF}\textbf{84.82±3.13\%} &
  59.86±1.81\% &
  59.43±2.68\% &
  60.81±1.87\% &
  64.75±1.76\% &
  -23.66\% \\ \midrule
Session 1 &
  24.24±3.91\% &
  26.13±7.62\% &
  51.02±1.70\% &
  \cellcolor[HTML]{EFEFEF}\textbf{61.09±1.41\%} &
  34.29±1.97\% &
  40.96±2.38\% &
  40.54±1.28\% &
  57.57±2.80\% &
  -5.76\% \\
Session 2 &
  17.54±4.98\% &
  14.83±0.92\% &
  37.77±4.29\% &
  45.53±3.08\% &
  28.30±4.16\% &
  38.78±1.97\% &
  37.03±5.06\% &
  \cellcolor[HTML]{EFEFEF}\textbf{50.11±2.03\%} &
  +10.05\% \\
Session 3 &
  16.13±5.55\% &
  8.55±0.71\% &
  32.79±2.65\% &
  33.73±1.49\% &
  23.37±3.63\% &
  35.43±5.63\% &
  32.93±7.96\% &
  \cellcolor[HTML]{EFEFEF}\textbf{45.21±1.04\%} &
  +27.60\% \\
Session 4 &
  8.41±2.11\% &
  6.30±2.31\% &
  24.39±2.54\% &
  31.63±2.35\% &
  23.77±5.02\% &
  38.16±5.31\% &
  31.27±6.79\% &
  \cellcolor[HTML]{EFEFEF}\textbf{41.77±0.79\%} &
  +9.46\% \\
Session 5 &
  9.04±5.46\% &
  4.94±2.48\% &
  22.01±1.28\% &
  28.15±1.17\% &
  20.23±4.00\% &
  32.74±4.57\% &
  26.57±5.83\% &
  \cellcolor[HTML]{EFEFEF}\textbf{36.26±2.79\%} &
  +10.75\% \\ \bottomrule
\end{tabular}%
}
\end{table*}

\subsection{Performance Comparison}

\begin{figure}
    \centering
    \includegraphics[width=\linewidth]{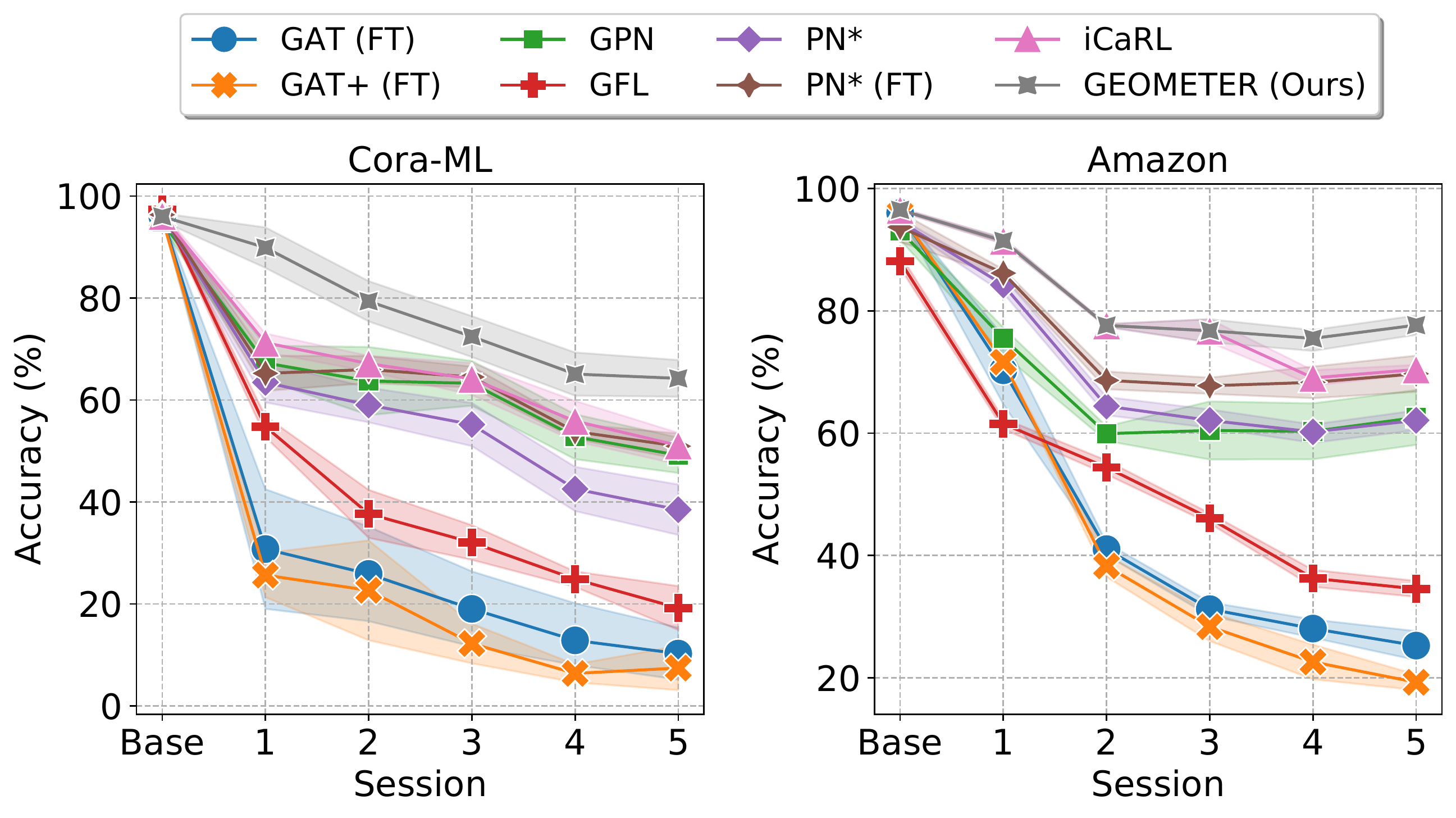}
    \caption{Comparison results of node classification accuracy in GFSCIL settings  on Cora-ML and Amazon datasets.}
    \label{fig:performance_two}
\end{figure}

We run \textsc{Geometer} and other baselines 10 times with different random seeds and report the average test accuracy over all encountered classes in Table \ref{tab:cora_flickr} and Figure \ref{fig:performance_two}. 
From the comprehensive views, we make the following observations:

(1) Firstly, our proposed model \textsc{Geometer} outperforms other baselines across Cora-ML, Amazon and Cora-Full datasets. For example, \textsc{Geometer} achieves 13.49\% to 24.15\% performance improvement over the best baseline model in 10 streaming sessions on Cora-Full dataset. Meanwhile, as shown in Figure \ref{fig:performance_two}, \textsc{Geometer} does not suffer dramatic performance degradation as other baselines, strongly demonstrating the superiority of our approach. 

(2) Secondly, by integrating the idea of metric learning, GFL, as the state-of-the-art model of graph few-shot learning, shows competitive performance at the base stage and first session on Flickr dataset. However, it is worth noting that \textsc{Geometer} achieves supreme results as more streaming session arrives and achieves substantial improvements. 

(3) GNN methods with fully connected neural network classifer largely fall behind other baselines. Those two methods need to replace and retrain the fully connected neural network classifer, which relies on sufficent training samples of each node classes. Therefore, with the increase of few-labeling novel classes, the performance of node classification deteriorates dramatically. PN* (FT) and iCaRL show better performance in several datasets, which shows metric-based methods with finetuning are more suitable for solving the GFSCIL problems.



\subsection{Ablation Study}

In this section, we analyze our \textsc{Geometer} model with several degenerate models from four aspects. Due to space limitations, the ablation studies are conducted on two representative datasets: Cora-ML and Amazon.

\begin{figure}
    \centering
    \includegraphics[width=\linewidth]{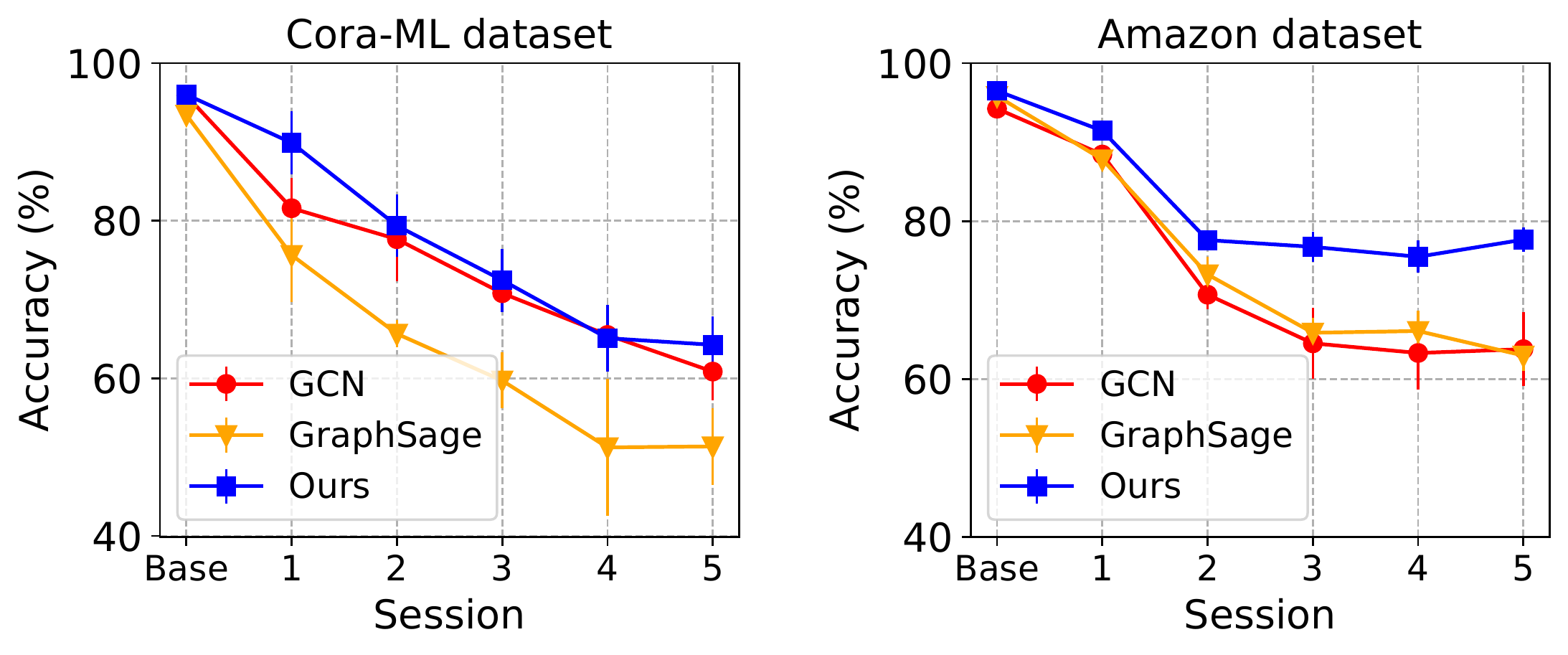}
    \caption{Ablation study of different GNN backbone on Cora-ML and Amazon dataset.}
    \label{fig:ablation_gnn}
\end{figure}

(1) \textbf{GNN backbone}: Due to the dynamic evolution of graph, \textsc{Geometer} utilizes graph attention network to capture the node features. In the ablation study, we replace it with two well-known GNN backbone GCN and GraphSage for comparison. As shown in Figure \ref{fig:ablation_gnn}, the performance of GCN and GraphSage falls behind graph attention network especially in Amazon dataset, since these two model fails to capture the complex correlations in message-passing.

\begin{table*}
\centering
\caption{Ablation study of loss functions comparsion on Cora-ML and Amazon dataset.}
\label{tab:loss_functions}
\resizebox{\linewidth}{!}{%
\begin{tabular}{@{}cccc|cccccccc@{}}
\toprule
\multicolumn{4}{c|}{\textbf{Loss functions}} &
  \multicolumn{4}{c}{\textbf{Cora-ML (1-way 5-shot GFSCIL setting)}} &
  \multicolumn{4}{c}{\textbf{Amazon (1-way 5-shot GFSCIL setting)}} \\ \midrule
\textbf{$\mathcal{L}_{P}$} &
  \textbf{$\mathcal{L}_{U}$} &
  \textbf{$\mathcal{L}_{S}$} &
  \textbf{$\mathcal{L}_{KD}$} &
  \textbf{Base Classes} &
  \textbf{Session 1} &
  \textbf{Session 3} &
  \textbf{Session 5} &
  \textbf{Base Classes} &
  \textbf{Session 1} &
  \textbf{Session 3} &
  \textbf{Session 5} \\ \midrule
$\checkmark$ &
  \multicolumn{1}{l}{} &
  \multicolumn{1}{l}{} &
  $\checkmark$ &
  \cellcolor[HTML]{EFEFEF}\textbf{96.21±0.67\%} &
  88.25±3.99\% &
  64.89±2.53\% &
  56.21±5.55\% &
  96.72±0.28\% &
  90.91±0.59\% &
  74.74±2.33\% &
  73.73±3.01\% \\
$\checkmark$ &
  $\checkmark$ &
  \multicolumn{1}{l}{} &
  $\checkmark$ &
  95.85±0.56\% &
  \cellcolor[HTML]{EFEFEF}\textbf{90.41±3.86\%} &
  68.26±3.65\% &
  58.72±4.66\% &
  96.72±0.22\% &
  91.39±0.56\% &
  76.55±1.94\% &
  73.97±1.90\% \\
$\checkmark$ &
  \multicolumn{1}{l}{} &
  $\checkmark$ &
  $\checkmark$ &
  95.71±0.55\% &
  89.21±2.88\% &
  69.57±2.71\% &
  54.28±3.90\% &
  96.83±0.32\% &
  91.15±0.35\% &
  75.08±2.61\% &
  73.92±2.60\% \\
$\checkmark$ &
  $\checkmark$ &
  $\checkmark$ &
   &
  95.74±0.61\% &
  90.40±4.82\% &
  68.16±1.45\% &
  62.41±2.37\% &
  \cellcolor[HTML]{EFEFEF}\textbf{96.86±0.35\%} &
  91.17±0.38\% &
  75.36±1.28\% &
  74.51±2.53\% \\ \midrule
$\checkmark$ &
  $\checkmark$ &
  $\checkmark$ &
  $\checkmark$ &
  96.01±0.92\% &
  89.89±3.97\% &
  \cellcolor[HTML]{EFEFEF}\textbf{72.45±4.01\%} &
  \cellcolor[HTML]{EFEFEF}\textbf{64.25±3.60\%} &
  96.50±0.29\% &
  \cellcolor[HTML]{EFEFEF}\textbf{91.44±0.46\%} &
  \cellcolor[HTML]{EFEFEF}\textbf{76.74±1.89\%} &
  \cellcolor[HTML]{EFEFEF}\textbf{77.66±1.58\%} \\ \bottomrule
\end{tabular}%
}
\end{table*}

\begin{figure}
    \centering
    \includegraphics[width=\linewidth]{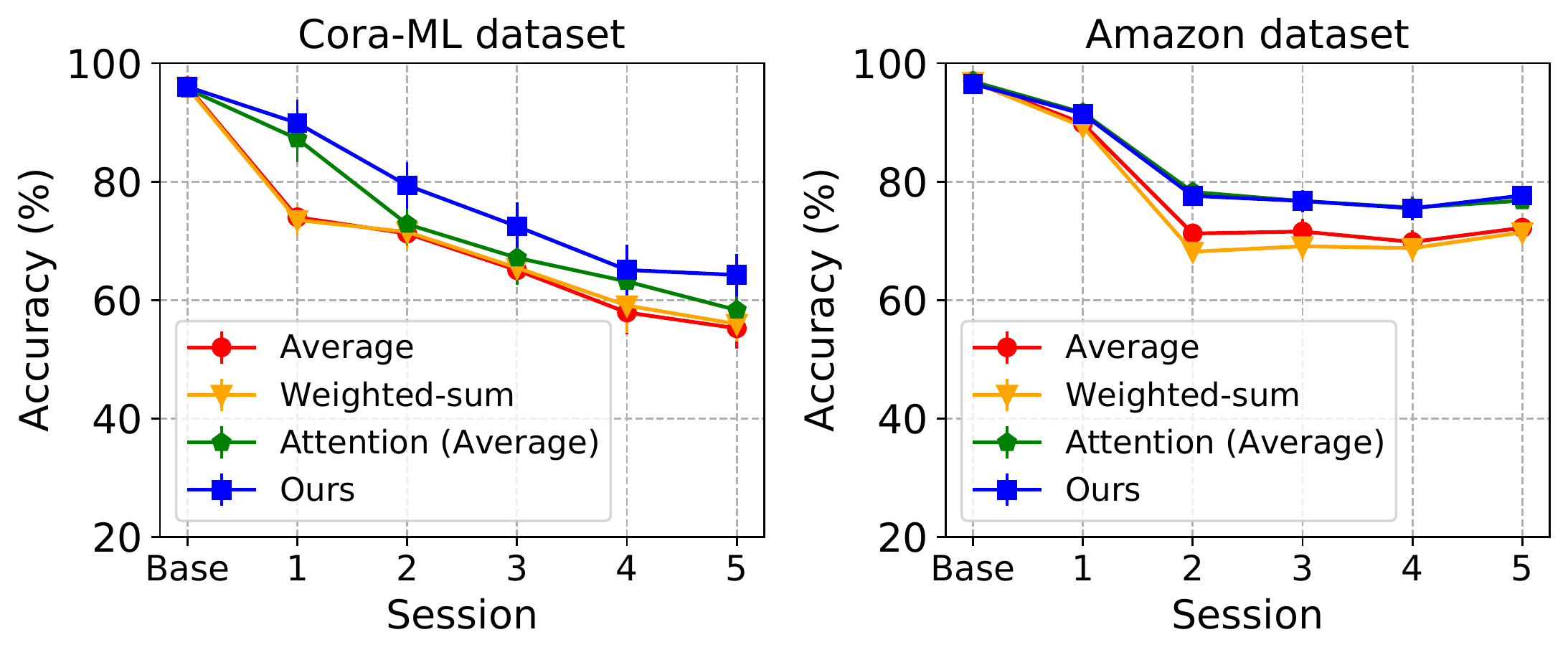}
    \caption{Ablation study of prototype representation methods on Cora-ML and Amazon dataset.}
    \label{fig:ablation_proto}
\end{figure}

(2) \textbf{Prototype representation method}: \textsc{Geometer} proposes class-level multi-head attention to learn the class prototype.
Three degenerate prototype representation methods are considered in ablation study-i.e. means of support node embedding (\text{Average}), weighted-sum of node embedding by node degree (\text{Weighted-sum}), and average node embeddings as the initial prototype in multi-head attention mechanism (\text{Attention (Average)}). Results on two datasets are presented in Figure \ref{fig:ablation_proto}. Since the unequal and non-static node influence, attention-based methods helps to better characterize the relationship of nodes and classes. On Amazon dataset, \textsc{Geometer} and \text{Attention (Average)} show close performance, while the introduction of degree-based weighted-sum further improve accuracy of prototype representation on Cora-ML dataset.

(3) \textbf{Loss functions}: \textsc{Geometer} proposes four different loss functions to finetune the prototype representation, and the effects of inter-class uniformity, inter-class separability and knowledge distillation are compared in Table \ref{tab:loss_functions}. In general, with the pop-up of novel classes, the best classification results are obtained when the four loss functions are combined together. Different loss functions optimize the classification effect from different aspects, and the deletion of any loss function will have a significant impact on performance.

\begin{figure}
    \centering
    \includegraphics[width=\linewidth]{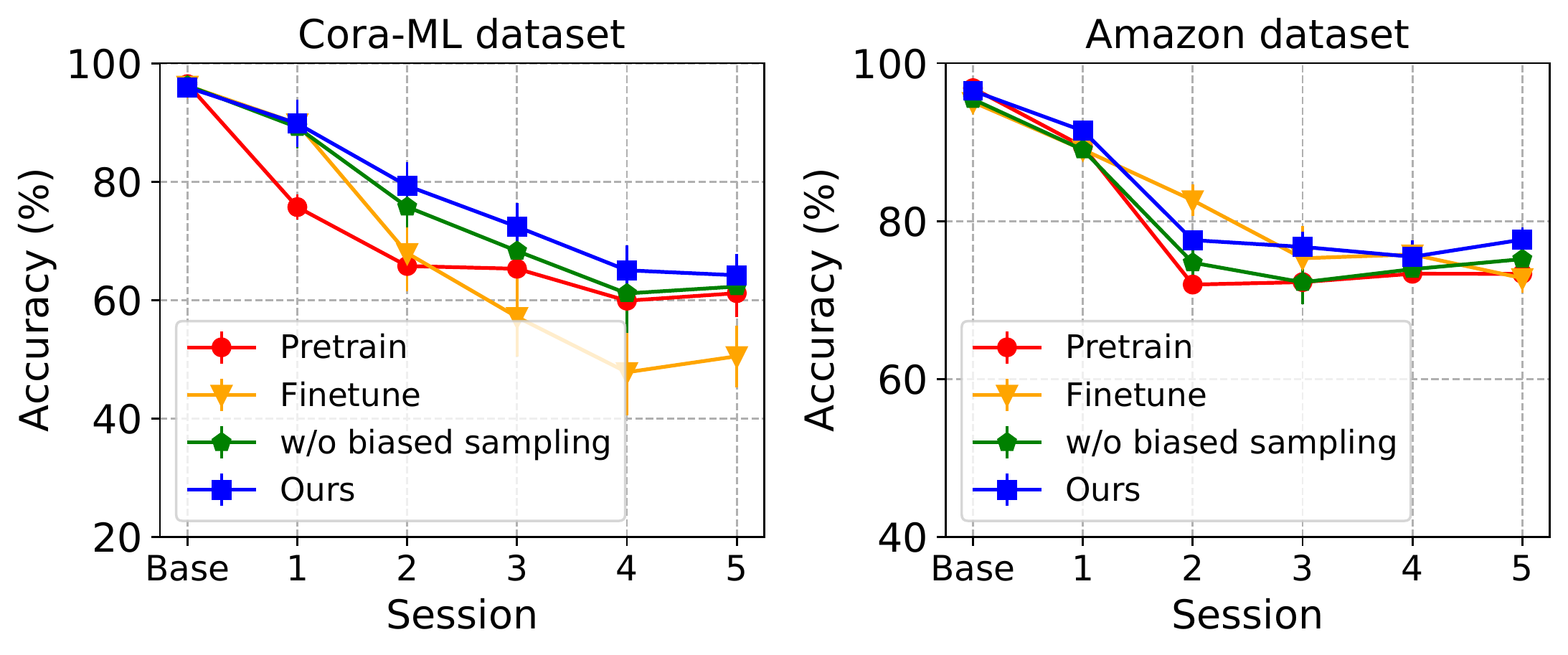}
    \caption{Ablation study of different biased sampling strategy on Cora-ML and Amazon dataset.}
    \label{fig:ablation_bias}
\end{figure}

(4) \textbf{Biased sampling strategy}: In episode meta learning, due to the imbalanced labeling of base and novel classes, different biased sampling strategies are adopted in pretraining stage and finetuning stage. In ablation study, we compare the performance with only one strategy is adopted or without biased sampling strategy in Figure \ref{fig:ablation_bias}. When we discard any biased sampling strategy, the learning process degenerates into a MAML-based learning strategy, and our method always shows better performance. If only one biased sampling strategy is adopted, partial sessions on Amazon datasets will have better results, but overall requires a combination of two biased sampling strategies.


\subsection{Parameter Analysis}

\begin{figure}
    \centering
    \includegraphics[width=\linewidth]{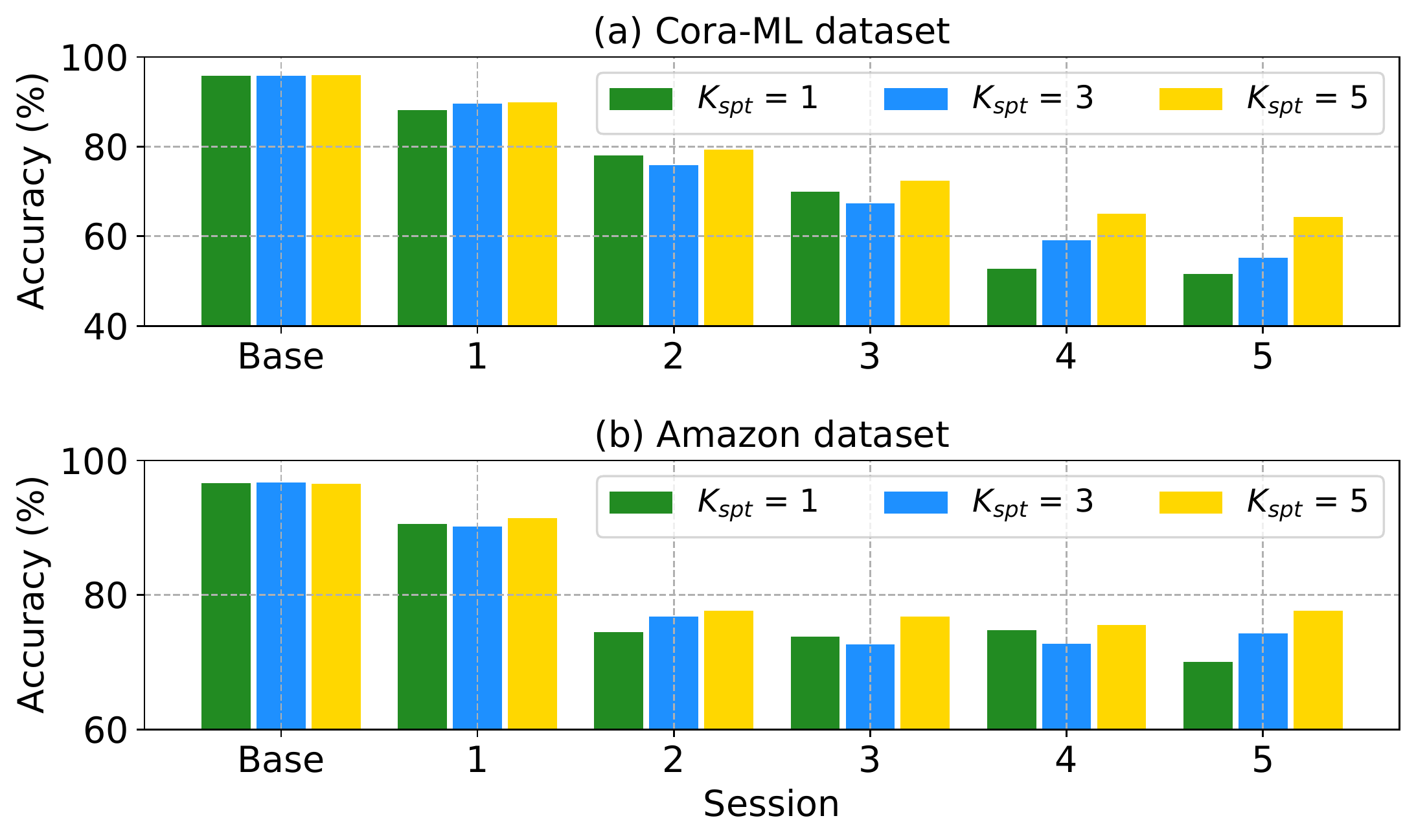}
    \caption{Parameter Analysis of support set size of novel classes on Cora-ML and Amazon dataset.}
    \label{fig:hyper_spt}
\end{figure}

In addition, we investigate the effect of support set size $K_{spt}$ on two dataset. By changing the the value of shot number $K_{spt}\in[1,3,5]$, we obtain different model performance. As shown in Figure \ref{fig:hyper_spt}, we can clearly observe that the performance of \textsc{Geometer} increases as the support set size $K_{spt}$, indicating that a larger support set helps to learn better class prototypes. At the same time, it shows that \textsc{Geometer} is able to overcome the noise or outliers due to few-shot labeling and effectively learn the class representations.

\subsection{Case Study}

\begin{figure}[h]
    \centering
    \includegraphics[width=\linewidth]{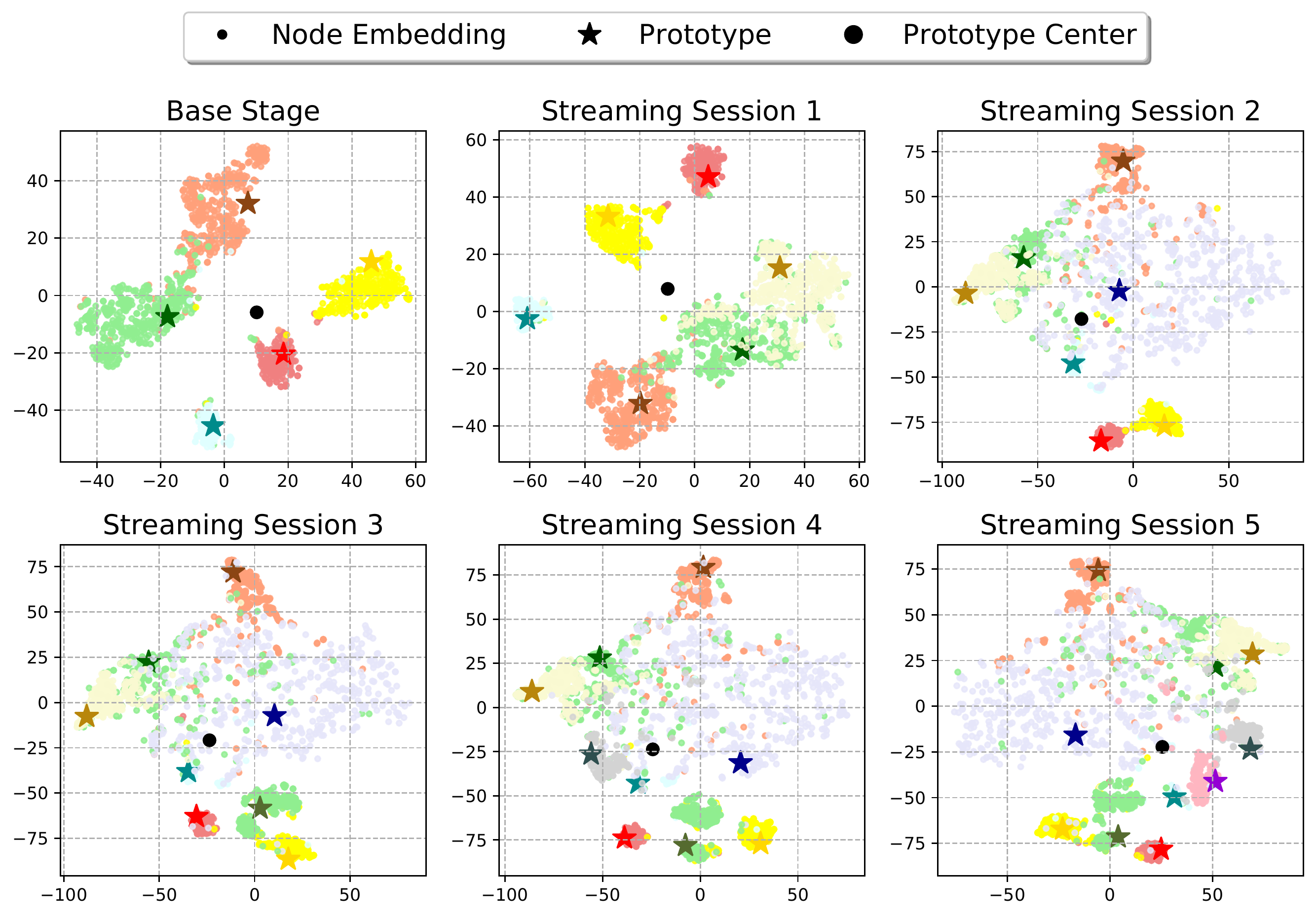}
    \caption{A $t$-SNE visualization of the query node embeddings and prototypes of \textsc{Geometer} (ours) on Amazon dataset.}
    \label{fig:tsne}
\end{figure}

In order to explore the effects of geometric losses and knowledge distillation technique, we use $t$-SNE method to project the node embeddings and prototype representations of base stage and 5 streaming sessions on Amazon dataset, as shown in Figure \ref{fig:tsne}. The visualization shows that as the novel classes arrival, most nodes are well clustered, and the prototypes are uniformly distributed around the prototype center. It is worth noting that a hard novel classes (colored in light purple) emerges in streaming session 2. Since \textsc{Geometer} takes into account the geometric relationships in the metric space and adapt knowledge distillation, the following novel classes in the subsequent streaming sessions actively distance with the light purple class, thereby avoiding the dramatic drop in performance caused by prototype representations overlapping.

\section{Conclusion}

In this paper, we propose \textsc{Geometer} for Graph Few-Shot Class-Incremental Learning (GFSCIL). As far as we known, this is the first work to deal with this challenging yet practical problem. The core idea of \textsc{Geometer} is to adjust the prototype representation in metric space from the aspects of geometric relationship and knowledge distillation, so as to realize the classification of ever-expanding classes with few-shot samples. Extensive experiments on four public datasets show that \textsc{Geometer} significantly outperforms the state-of-the-art baselines. In the future, we would like to extend our framework to address more challenging problem, like the open-set classification in graphs.

\section*{Acknowledgement}
This work was supported by Natural Science Foundation of China under Grants No. 42050105; in part by NSF China (No. 62020106005, 62061146002, 61960206002, 61829201, 61832013), 2021 Tencent AI Lab RhinoBird Focused Research Program (No: JR202132), and the Program of Shanghai Academic/Technology Research Leader under Grant No. 18XD1401800.

\bibliographystyle{unsrt}
\bibliography{reference}

\begin{thebibliography}{10}

\bibitem{wang2020gcn}
Xiao Wang, Meiqi Zhu, Deyu Bo, Peng Cui, Chuan Shi, and Jian Pei.
\newblock Am-gcn: Adaptive multi-channel graph convolutional networks.
\newblock In {\em Proceedings of the 26th ACM SIGKDD International conference
  on knowledge discovery \& data mining}, pages 1243--1253, 2020.

\bibitem{xhonneux2020continuous}
Louis-Pascal Xhonneux, Meng Qu, and Jian Tang.
\newblock Continuous graph neural networks.
\newblock In {\em International Conference on Machine Learning}, pages
  10432--10441. PMLR, 2020.

\bibitem{DBLP:conf/kdd/WangWGG21}
Zheng Wang, Jialong Wang, Yuchen Guo, and Zhiguo Gong.
\newblock Zero-shot node classification with decomposed graph prototype
  network.
\newblock In {\em Proccedings of The 27th {ACM} {SIGKDD} Conference on
  Knowledge Discovery and Data Mining, Virtual Event, Singapore, August 14-18,
  2021}, pages 1769--1779. {ACM}, 2021.

\bibitem{DBLP:conf/kdd/QiuCDZYDWT20}
Jiezhong Qiu, Qibin Chen, Yuxiao Dong, Jing Zhang, Hongxia Yang, Ming Ding,
  Kuansan Wang, and Jie Tang.
\newblock {GCC:} graph contrastive coding for graph neural network
  pre-training.
\newblock In {\em The 26th {ACM} {SIGKDD} Conference on Knowledge Discovery and
  Data Mining, Virtual Event, CA, USA, August 23-27, 2020}, pages 1150--1160.
  {ACM}, 2020.

\bibitem{DBLP:conf/aaai/YouGYL21}
Jiaxuan You, Jonathan~M. Gomes{-}Selman, Rex Ying, and Jure Leskovec.
\newblock Identity-aware graph neural networks.
\newblock In {\em Thirty-Fifth {AAAI} Conference on Artificial Intelligence,
  Virtual Event, February 2-9, 2021}, pages 10737--10745, 2021.

\bibitem{DBLP:conf/iclr/KipfW17}
Thomas~N. Kipf and Max Welling.
\newblock Semi-supervised classification with graph convolutional networks.
\newblock In {\em International Conference on Learning Representations}, 2017.

\bibitem{DBLP:conf/nips/Huang0RH18}
Wen{-}bing Huang, Tong Zhang, Yu~Rong, and Junzhou Huang.
\newblock Adaptive sampling towards fast graph representation learning.
\newblock In {\em Advances in Neural Information Processing Systems}, pages
  4563--4572, 2018.

\bibitem{DBLP:conf/aaai/BoWSS21}
Deyu Bo, Xiao Wang, Chuan Shi, and Huawei Shen.
\newblock Beyond low-frequency information in graph convolutional networks.
\newblock In {\em Thirty-Fifth {AAAI} Conference on Artificial Intelligence,
  Virtual Event, February 2-9, 2021}, pages 3950--3957, 2021.

\bibitem{zhou2019meta}
Fan Zhou, Chengtai Cao, Kunpeng Zhang, Goce Trajcevski, Ting Zhong, and
  Ji~Geng.
\newblock Meta-gnn: On few-shot node classification in graph meta-learning.
\newblock In {\em Proceedings of the 28th ACM International Conference on
  Information and Knowledge Management}, pages 2357--2360, 2019.

\bibitem{DBLP:conf/nips/HuangZ20}
Kexin Huang and Marinka Zitnik.
\newblock Graph meta learning via local subgraphs.
\newblock In {\em Advances in Neural Information Processing Systems}, 2020.

\bibitem{liu2021relative}
Zemin Liu, Yuan Fang, Chenghao Liu, and Steven~CH Hoi.
\newblock Relative and absolute location embedding for few-shot node
  classification on graph.
\newblock In {\em Proceedings of the AAAI Conference on Artificial
  Intelligence}, volume~35, pages 4267--4275, 2021.

\bibitem{li2017learning}
Zhizhong Li and Derek Hoiem.
\newblock Learning without forgetting.
\newblock {\em IEEE Transactions on Pattern Analysis and Machine Intelligence},
  40(12):2935--2947, 2017.

\bibitem{rebuffi2017icarl}
Sylvestre-Alvise Rebuffi, Alexander Kolesnikov, Georg Sperl, and Christoph~H
  Lampert.
\newblock icarl: Incremental classifier and representation learning.
\newblock In {\em Proceedings of the IEEE conference on Computer Vision and
  Pattern Recognition}, pages 2001--2010, 2017.

\bibitem{hou2019learning}
Saihui Hou, Xinyu Pan, Chen~Change Loy, Zilei Wang, and Dahua Lin.
\newblock Learning a unified classifier incrementally via rebalancing.
\newblock In {\em Proceedings of the IEEE/CVF Conference on Computer Vision and
  Pattern Recognition}, pages 831--839, 2019.

\bibitem{DBLP:journals/corr/GoodfellowMDCB13}
Ian~J. Goodfellow, Mehdi Mirza, Xia Da, Aaron~C. Courville, and Yoshua Bengio.
\newblock An empirical investigation of catastrophic forgeting in
  gradient-based neural networks.
\newblock In {\em 2nd International Conference on Learning Representations},
  2014.

\bibitem{kirkpatrick2017overcoming}
James Kirkpatrick, Razvan Pascanu, Neil Rabinowitz, Joel Veness, Guillaume
  Desjardins, Andrei~A Rusu, Kieran Milan, John Quan, Tiago Ramalho, Agnieszka
  Grabska-Barwinska, et~al.
\newblock Overcoming catastrophic forgetting in neural networks.
\newblock {\em Proceedings of the National Academy of Sciences},
  114(13):3521--3526, 2017.

\bibitem{DBLP:conf/icml/YapRB21}
Pau~Ching Yap, Hippolyt Ritter, and David Barber.
\newblock Addressing catastrophic forgetting in few-shot problems.
\newblock In {\em Proceedings of the 38th International Conference on Machine
  Learning, {ICML} 2021, 18-24 July 2021, Virtual Event}, volume 139 of {\em
  Proceedings of Machine Learning Research}, pages 11909--11919. {PMLR}, 2021.

\bibitem{finn2017model}
Chelsea Finn, Pieter Abbeel, and Sergey Levine.
\newblock Model-agnostic meta-learning for fast adaptation of deep networks.
\newblock In {\em International Conference on Machine Learning}, pages
  1126--1135. PMLR, 2017.

\bibitem{ding2020graph}
Kaize Ding, Jianling Wang, Jundong Li, Kai Shu, Chenghao Liu, and Huan Liu.
\newblock Graph prototypical networks for few-shot learning on attributed
  networks.
\newblock In {\em Proceedings of the 29th ACM International Conference on
  Information \& Knowledge Management}, pages 295--304, 2020.

\bibitem{yao2020graph}
Huaxiu Yao, Chuxu Zhang, Ying Wei, Meng Jiang, Suhang Wang, Junzhou Huang,
  Nitesh Chawla, and Zhenhui Li.
\newblock Graph few-shot learning via knowledge transfer.
\newblock In {\em Proceedings of the AAAI Conference on Artificial
  Intelligence}, volume~34, pages 6656--6663, 2020.

\bibitem{castro2018end}
Francisco~M Castro, Manuel~J Mar{\'\i}n-Jim{\'e}nez, Nicol{\'a}s Guil, Cordelia
  Schmid, and Karteek Alahari.
\newblock End-to-end incremental learning.
\newblock In {\em Proceedings of the European conference on computer vision
  (ECCV)}, pages 233--248, 2018.

\bibitem{tao2020few}
Xiaoyu Tao, Xiaopeng Hong, Xinyuan Chang, Songlin Dong, Xing Wei, and Yihong
  Gong.
\newblock Few-shot class-incremental learning.
\newblock In {\em Proceedings of the IEEE/CVF Conference on Computer Vision and
  Pattern Recognition}, pages 12183--12192, 2020.

\bibitem{cheraghian2021semantic}
Ali Cheraghian, Shafin Rahman, Pengfei Fang, Soumava~Kumar Roy, Lars Petersson,
  and Mehrtash Harandi.
\newblock Semantic-aware knowledge distillation for few-shot class-incremental
  learning.
\newblock In {\em Proceedings of the IEEE/CVF Conference on Computer Vision and
  Pattern Recognition}, pages 2534--2543, 2021.

\bibitem{snell2017prototypical}
Jake Snell, Kevin Swersky, and Richard Zemel.
\newblock Prototypical networks for few-shot learning.
\newblock In {\em Proceedings of the 31st International Conference on Neural
  Information Processing Systems}, pages 4080--4090, 2017.

\bibitem{DBLP:conf/iclr/VelickovicCCRLB18}
Petar Velickovic, Guillem Cucurull, Arantxa Casanova, Adriana Romero, Pietro
  Li{\`{o}}, and Yoshua Bengio.
\newblock Graph attention networks.
\newblock In {\em International Conference on Learning Representations}, 2018.

\bibitem{vaswani2017attention}
Ashish Vaswani, Noam Shazeer, Niki Parmar, Jakob Uszkoreit, Llion Jones,
  Aidan~N Gomez, {\L}ukasz Kaiser, and Illia Polosukhin.
\newblock Attention is all you need.
\newblock In {\em Advances in Neural Information Processing Systems}, pages
  5998--6008, 2017.

\bibitem{hinton2015distilling}
Geoffrey Hinton, Oriol Vinyals, and Jeff Dean.
\newblock Distilling the knowledge in a neural network.
\newblock {\em arXiv preprint arXiv:1503.02531}, 2015.

\bibitem{10.1145/3447548.3467319}
SeongKu Kang, Junyoung Hwang, Wonbin Kweon, and Hwanjo Yu.
\newblock Topology distillation for recommender system.
\newblock In {\em Proccedings of The 27th {ACM} {SIGKDD} Conference on
  Knowledge Discovery and Data Mining, Virtual Event, Singapore, August 14-18,
  2021}, page 829–839, New York, NY, USA, 2021. {ACM}.

\bibitem{10.1145/3394486.3403234}
Qi~Wang, Liang Zhan, Paul Thompson, and Jiayu Zhou.
\newblock Multimodal learning with incomplete modalities by knowledge
  distillation.
\newblock In {\em Proceedings of the 26th ACM SIGKDD International Conference
  on Knowledge Discovery and Data Mining}, page 1828–1838, New York, NY, USA,
  2020. {ACM}.

\bibitem{bojchevski2018deep}
Aleksandar Bojchevski and Stephan G{\"u}nnemann.
\newblock Deep gaussian embedding of graphs: Unsupervised inductive learning
  via ranking.
\newblock In {\em International Conference on Learning Representations}, pages
  1--13, 2018.

\bibitem{graphsaint-iclr20}
Hanqing Zeng, Hongkuan Zhou, Ajitesh Srivastava, Rajgopal Kannan, and Viktor
  Prasanna.
\newblock {GraphSAINT}: Graph sampling based inductive learning method.
\newblock In {\em International Conference on Learning Representations}, 2020.

\bibitem{Hou2020Measuring}
Yifan Hou, Jian Zhang, James Cheng, Kaili Ma, Richard T.~B. Ma, Hongzhi Chen,
  and Ming-Chang Yang.
\newblock Measuring and improving the use of graph information in graph neural
  networks.
\newblock In {\em International Conference on Learning Representations}, 2020.

\end{thebibliography}

\appendix

\section{Appendix}

To support the reproducibility of the results in this paper, we have released our code and data. We implement the \textsc{Geometer} model based on Pytorch framework.\footnote{The implementation code and details of our model is available at https://github.com/RobinLu1209/Geometer.} 
All the evaluated models are implemented on a server with two CPUs (Intel Xeon E5-2630 $\times$ 2) and four GPUs (NVIDIA GTX 2080 $\times$ 4). 

\subsection{Dataset}
\label{appendix-dataset}
In this paper, we evaluate the proposed \textsc{Geometer} on four public datasets as follows:
\begin{itemize}[left=1em]
    \item \textbf{{Cora-ML}}~\cite{bojchevski2018deep} is an academic network about machine learning papers. The dataset contains 7 different classes, in which each node represents a paper and each edge represents the citation relationship between two papers.
    \item \textbf{{Flickr}}~\cite{graphsaint-iclr20} is a photo-sharing social network from Flickr.
     Each node represents one picture uploaded to the Flickr website and the node feature contains information of low-level feature from NUS-WIDE Dataset. Flickr forms the edges between images from the same location, submitted to the same gallery, sharing common tags, taken by friends, etc. 
    \item \textbf{{Amazon}}~\cite{Hou2020Measuring} is the segments of Amazon co-purchase e-commerce network, in which each node is an item and each edge denotes the co-purchasing relationship by a common user. The node features are bag-of-words encoded product reviews, and class labels are given by the product category.
    \item \textbf{{Cora-Full}}~\cite{bojchevski2018deep} is a well-known citation network labeled based on the paper topic, which has 70 different classes of papers. Among the academic networks we know, it has the largest network size and the largest number of categories.
\end{itemize}

\end{sloppypar}
\end{document}